\pdfoutput=1

\documentclass[11pt]{article}

\usepackage{EACL2023}

\usepackage{times}
\usepackage{latexsym}

\usepackage[T1]{fontenc}

\usepackage[utf8]{inputenc}

\usepackage{microtype}

\usepackage{inconsolata}
\usepackage{subcaption}
\usepackage{graphicx}
\usepackage{multirow}
\usepackage{hyperref}
\usepackage{paralist}
\usepackage{enumitem}

\usepackage{titlesec}
\titlespacing{\paragraph}{%
  0pt}{
  0.2\baselineskip}{
  1em}%
  
\title{BanglaNLG and BanglaT5: Benchmarks and Resources for \\ Evaluating Low-Resource Natural Language Generation in Bangla}

\author{
Abhik Bhattacharjee$^1$, Tahmid Hasan$^1$, Wasi Uddin Ahmad$^2$, Rifat Shahriyar$^1$\\ [3pt]
Bangladesh University of Engineering and Technology (BUET)$^1$, \\ University of California, Los Angeles$^2$\\[3pt]
\texttt{abhik@ra.cse.buet.ac.bd}, \texttt{\{tahmidhasan,rifat\}@cse.buet.ac.bd}\\
}

\begin{document}
\maketitle
\begin{abstract}

This work presents `BanglaNLG,' a comprehensive benchmark for evaluating natural language generation (NLG) models in Bangla, a widely spoken yet low-resource language. We aggregate six challenging conditional text generation tasks under the BanglaNLG benchmark, introducing a new dataset on dialogue generation in the process. Furthermore, using a clean corpus of 27.5 GB of Bangla data, we pretrain `BanglaT5', a sequence-to-sequence Transformer language model for Bangla. BanglaT5 achieves state-of-the-art performance in all of these tasks, outperforming several multilingual models by up to 9\% absolute gain and 32\% relative gain. We are making the new dialogue dataset and the BanglaT5 model publicly available at \url{https://github.com/csebuetnlp/BanglaNLG} in the hope of advancing future research on Bangla NLG.

\end{abstract}

\section{Introduction}

The emergence of pretrained language models \citep{devlin-etal-2019-bert, radford2019language, liu2019roberta} has brought about a revolutionary change in natural language processing (NLP). With little task-specific fine-tuning, these models have achieved state-of-the-art results on many NLP tasks \citep{wang2018glue, rajpurkar2016squad, sang2003introduction}. However, the focus of these models has predominantly been on natural language understanding (NLU). Even models pretrained with generative objectives \citep{raffel2019exploring} concern themselves with NLU tasks more than natural language generation (NLG) tasks. Although there have been recent efforts to uplift NLG \cite{gehrmann-etal-2021-gem}, they are primarily geared towards high- and mid-resource languages. For example, despite being the sixth most spoken language in the world with over 230 million native speakers comprising 3\% of the world's total population,\footnote{\url{https://w.wiki/Psq}} Bangla has remained an underrepresented language in the NLP literature \citep{joshi-etal-2020-state}. There have been only a handful of benchmark studies on Bangla NLG \citep{dabre-etal-2022-indicbart, kumar2022indicnlg}, and that too without Bangla being the main focus. This can be attributed to the lack of diverse NLG tasks under a single benchmark and strong pretrained Bangla NLG models. 

To this end, we present `\textbf{BanglaNLG},' a comprehensive benchmark for Bangla language generation comprising six representative tasks on machine translation, text summarization, question answering, dialogue generation, headline generation, and cross-lingual summarization. To our knowledge, BanglaNLG is the first NLG benchmark exclusively for a low-resource language. 

To establish a strong baseline for this benchmark, we pretrain \textbf{BanglaT5} -- a sequence-to-sequence Transformer model \citep{vaswani2017attention} pretrained on a 27.5 GB clean Bangla text corpus covering a broad range of domains. In summary:
\begin{itemize}
    \setlength\itemsep{-1mm}
    \item We develop the BanglaNLG benchmark bringing together six NLG tasks.
    \item We introduce a Multi-turn Dialogue dataset.
    \item We pretrain BanglaT5 and evaluate it on the six NLG tasks, showing strong results.
\end{itemize}

BanglaT5 outperforms similar-sized multilingual models, achieving new state-of-the-art results on three tasks with a 4\% gain on average. We are releasing the BanglaT5 model and a live leaderboard to promote future research on Bangla NLG.

\section{The Bangla Natural Language Generation (BanglaNLG) Benchmark}

\begin{table*}[!t]
\centering
\resizebox{\linewidth}{!}
{%
\begin{tabular}{llrrrll}
\hline
\textbf{Task} & \textbf{Corpus} & \textbf{|Train|} & \textbf{|Dev|} & \textbf{|Test|} & \textbf{Metric} & \textbf{Domain}\\
\hline
Machine Translation & BanglaNMT, FLoRes & 2,751,315 & 997 & 1,012 & SacreBLEU & Misc.\\ 
Text Summarization & XL-Sum & 8,102 & 1,012 & 1,012 & ROUGE-2 & BBC\\
Question Answering & BQA & 127,771 & 2,502 & 2,504 & EM/F1 & Wikipedia\\
Multi-turn Dialogue & DailyDialog & 76,052 & 7,069 & 6,640 & BLEU-1 & Misc.\\
News Headline Generation & XL-Sum & 8,102 & 1,012 & 1,012 & ROUGE-2 & BBC\\
Cross-lingual Summarization & CrossSum & 1241 & 153 & 155 & ROUGE-2 & BBC\\
\hline
\end{tabular}
}
\caption{
Dataset statistics and basic characteristics of BanglaNLG. Machine translation and cross-lingual summarization datasets include examples of Bangla $\leftrightarrow$ English.
}
\label{tab:blub}
\end{table*}

There have been sporadic works on Bangla NLG, mostly catered to machine translation \cite{hasan-etal-2020-low,10.3844/jcssp.2019.1627.1637, 10.3844/jcssp.2019.1022.1039} and text summarization \cite{10.1007/978-981-33-4673-4_4, 9689852}. However, Bangla NLG lacks a unified study comprising diverse and challenging tasks. Motivated by the popular benchmarks like GLUE \citep{wang2018glue}, XTREME \citep{pmlr-v119-hu20b}, GEM \citep{gehrmann-etal-2021-gem}, that have facilitated the training/evaluation of NLP models, we establish the first-ever Bangla Natural Language Generation (BanglaNLG) Benchmark. 

\subsection{Task Selection Criteria}

We consider the following factors while choosing the evaluation tasks:
\paragraph{1. Diversity:} The tasks should focus on evaluating the model's generalization capabilities. Therefore, they should vary in task nature -- the input and output length, the type of generated text, the target domain, and the dataset size.
\paragraph{2. Practical Applicability:} The choice of tasks should be driven by their practical implications. Rather than being used in abstract situations, NLG models trained on these tasks should be able to aid/reduce human effort in real-world scenarios.
\paragraph{3. Difficulty:} The tasks should be challenging while not being unsolvable. There should be clear room for improvement to foster future research.
\paragraph{4. Accessibility:} The selected datasets for these tasks should be openly accessible to encourage researchers to design better NLG models.
\paragraph{5. Evaluation:} The selected tasks should have reliable automated metrics for evaluating the focused abilities of an NLG model.


\subsection{Selected Tasks}

Considering the criteria mentioned above, we design BanglaNLG as an aggregation of six tasks:

\paragraph{1. Machine Translation (MT):} MT is perhaps the most studied NLG task in Bangla and the most commonly benchmarked NLG task in general. We use the BanglaNMT parallel corpus \citep{hasan-etal-2020-low}, the largest Bangla-English MT dataset curated, with 2.75 million parallel pairs for training. The sentence pairs originate from various domains such as Wikipedia, news articles, religious and law documents, etc. We evaluate the NLG models using FLoRes-100 \citep{goyal-etal-2022-flores} in both directions on this dataset, i.e., Bangla to English and English to Bangla. This task is particularly challenging since it assesses an NLG model's bilingual generation capabilities. Following standard practice, we use detokenized SacreBLEU \cite{post-2018-call} as the evaluation metric for this task.

\paragraph{2. Text Summarization (TS):} This task aims to generate a  short and fluent summary given a long text document. We chose the Bangla portion of XL-Sum \citep{hasan-etal-2021-xl} for this task. XL-Sum is a large comprehensive dataset for abstractive TS where the article and summaries are written by professional editors of BBC News. The articles cover various topics such as entertainment, politics, science, sports, etc. For this task, we use ROUGE-2\footnote{We use Bangla stemming supported ROUGE implementation from \url{https://github.com/csebuetnlp/xl-sum/tree/master/multilingual_rouge_scoring}.} \cite{lin-2004-rouge} as the evaluation metric.  

\paragraph{3. Question Answering (QA):} This is a fundamental NLP task that can be modeled as both an NLU and NLG task. We use the BQA \cite{bhattacharjee-etal-2022-banglabert} dataset for this task. The training data is machine translated from SQuAD 2.0 \cite{rajpurkar-etal-2018-know}, while the evaluation data come from the human-annotated question-answer pairs of the TyDi-QA \cite{clark-etal-2020-tydi} secondary gold passage task. Although TyDi-QA only contains answerable questions, BQA introduced unanswerable questions to make the task more challenging. Following SQuAD 2.0, we use Exact Match (EM) and F1 as the evaluation metrics.

\paragraph{4. Multi-turn Dialogue (MTD):} Conversational AI is a crucial task for NLG \cite{chen2017survey}. However, there is no public dataset for dialogue generation in Bangla. As such, we curate a new multi-turn dialogue dataset by translating the DailyDialog \citep{li-etal-2017-dailydialog} dataset using the English to Bangla translation model introduced by \citet{hasan-etal-2020-low}. Unlike standard QA-style conversation datasets, DailyDialog reflects real-life conversations in various social situations rich in emotion, making it a perfect candidate for our benchmark. We automatically translate the training data following the same procedure described in \citet{bhattacharjee-etal-2022-banglabert} and have the evaluation sets manually translated by expert human translators. We use BLEU-1 as the evaluation metric for this task to properly differentiate between models since averaged BLEU scores of up to 4-gram tend to be quite low in dialogue evaluation \cite{zhang-etal-2020-dialogpt}.

\paragraph{5. News Headline Generation (NHG):} Automating headline generation can help news editors write compelling headlines to draw readers' attention. We consider NHG as a complementary task to TS. Given an article, the objective is to generate an appropriate headline that accurately depicts the article. We repurpose the XL-Sum \citep{hasan-etal-2021-xl} dataset for this task since it also includes the titles of the articles. Like TS, we use ROUGE-2 as the evaluation metric.

\paragraph{6. Cross-lingual Summarization (XLS):} As another task for evaluating models' bilingual generation capabilities, we consider XLS. In this task, given a piece of text in a source language, we have to generate the corresponding summary in a target language. This is potentially harder than both MT and TS considering it combines both in a single task. We consider the English-Bengali portion of the CrossSum \cite{bhattacharjee2021crosssum} dataset for this task. It is curated by aligning identical articles written in different languages from the XL-Sum dataset. For evaluation, we use ROUGE-2. 

We present detailed statistics of the BanglaNLG benchmark in Table \ref{tab:blub}.
\begin{table*}[!tbh]
\centering \setlength{\tabcolsep}{7pt}
\begin{tabular}{lc cccccc}
\hline
\textbf{Model} & \textbf{Parameters} & \textbf{MT} & \textbf{TS} & \textbf{QA} & \textbf{MTD} & \textbf{NHG} & \textbf{XLS}\\
\hline
mT5 (base) & 582M & 30.1/\textbf{17.2} &  10.3 & 59.0/65.3 & 17.5 & 9.6 & 2.7/0.7\\
XLM-ProphetNet & 616M &  27.5/15.4 & 7.8 & 53.0/57.3 & \textbf{20.0} & 9.5 & \textbf{6.2}/2.7\\
mBART-50 & 611M &  29.7/15.5 & 10.4 & 53.4/58.9 & \textbf{18.5} & 11.2 & 5.4/\textbf{3.7}\\
IndicBART (unified) & 244M &  28.1/16.6 & 8.9 & 59.6/65.6 & 14.8 & 7.9 & \textbf{6.3}/2.5\\
IndicBART (separate) & 244M &  27.5/15.7 & 9.2 & 55.3/61.2 & 14.1 & 9.1 & 5.3/2.4\\
BanglaT5 & 247M & \textbf{31.3}/\textbf{17.4} &  \textbf{13.7} & \textbf{68.5}/\textbf{74.8} & \textbf{19.0} & \textbf{13.8} & \textbf{6.4}/\textbf{4.0}\\
\hline
\end{tabular}
\caption{Performance comparison of the pretrained models on different BanglaNLG tasks. Scores in bold texts have statistically significant ($p < 0.05$) difference from others with bootstrap sampling \citep{koehn-2004-statistical}.}\label{tab:benchmark}
\end{table*}

\section{BanglaT5}\label{sec:pretraining}

We introduce BanglaT5, a sequence-to-sequence
Transformer model \cite{vaswani2017attention}, to establish a strong baseline for BanglaNLG benchmark.
In this section, we describe the pretraining data, objectives, and model architecture of BanglaT5.

\subsection{Pretraining Data}

We chose Bangla2B+ \citep{bhattacharjee-etal-2022-banglabert} as the pretraining corpus for BanglaT5. This is a 27.5 GB dataset containing 5.25 million documents collected from a meticulously selected list of web sources. While larger sources like CCNet \cite{wenzek-etal-2020-ccnet} and mC4 \citep{xue-etal-2021-mt5} are available, these contain a lot of noise and offensive texts that are difficult to remove. For a generative model, even small amounts of unwanted texts in pretraining could lead to potentially dangerous biases in generated text \cite{luccioni-viviano-2021-whats}. Therefore, we decided not to use them. 

\subsection{Data Pre-processing}

Following \citet{hasan-etal-2020-low}, we preprocessed the texts using their normalization pipeline\footnote{\url{https://github.com/csebuetnlp/normalizer}}. We trained a SentencePiece \citep{kudo-richardson-2018-sentencepiece} vocabulary of 32k subword tokens on the normalized corpus with a character coverage of 0.99995. While creating a training sample, we limited the maximum sequence length to 512 tokens for both input and output and discarded documents with a token count below 7. After tokenization, we had 4.8 million data points with an average sequence length of 402.32 tokens. 

\subsection{Pretraining Objective}

For generative language modeling, two standard choices are decoder-only models \citep{mikolov2010recurrent} and encoder-decoder models \citep{sutskever2014sequence}. \citet{radford2019language}  trained a decoder-only Transformer \cite{vaswani2017attention} pretrained on the conditional continuation objective. However, to provide more flexibility on generation and possible usage on understanding tasks, we only consider encoder-decoder models following the original design of the Transformer. They are generally trained with different denoising objectives to increase the encoder's and decoder's capacity. For instance, BART \cite{lewis-etal-2020-bart}, and mBART \cite{liu-etal-2020-multilingual-denoising} use a text-infilling-based objective. In contrast, MARGE \cite{10.5555/3495724.3497275} is a multilingual encoder-decoder model trained to reconstruct a document in one language by retrieving documents in other languages. Following \citet{raffel2019exploring}, we pretrained BanglaT5 using a "span-correction" objective, empirically shown to be an optimal choice for encoder-decoder models. In this objective, consecutive spans of input tokens are replaced with a mask token, and the model is trained to reconstruct them. 
 
\subsection{Model Architecture \& Hyperparameters}

We pretrained the base variant of the T5 model: 12 layers, 12 attention heads, 768 hidden size, 2048 feed-forward size with GeGLU activation \cite{shazeer2020glu} with a batch size of 65536 tokens for 3 million steps on a v3-8 TPU instance on GCP. We used the Adam \cite{kingma2014adam} optimizer with a 3e-4 learning rate, linear warmup of 10k steps, and `inverse square root' learning rate decay.
\section{Experiments \& Results}
\label{sec:experiments}

We compared BanglaT5 it with four multilingual models: mT5 (base) \cite{xue-etal-2021-mt5}, mBART-50 \citep{tang2020multilingual}, XLM-ProphetNet \citep{qi-etal-2021-prophetnet}, and IndicBART (both unified and separate script variants) \citep{dabre-etal-2022-indicbart}.\footnote{Due to computational budget limitations, we do not benchmark on billion-parameter models like large mT5 variants.} 
All pretrained models were fine-tuned for 3-15 epochs with batch size 32 (128 for MT). We used linear warmup with a ratio of 0.1, label smoothing of 0.1 \citep{7780677}, and weight decay of 1e-6 with the Adam optimizer \citep{kingma2014adam}. The learning rate was tuned from the set \{5e-5, 1e-4, 5e-4\}. The best model was evaluated based on the validation performance after each epoch.

During inference, we used beam-search \cite{hayes1976speech} with beam size 5 (on all tasks except QA), removed duplicated trigrams during beam search \cite{fan-etal-2018-controllable}, and used a length penalty \cite{DBLP:journals/corr/WuSCLNMKCGMKSJL16} of 0.6. For QA, we used greedy decoding, i.e., picking the most probable token during each decoding step.

The evaluation results are presented in Table \ref{tab:benchmark}. In all the tasks, BanglaT5 outperformed all multilingual models by a considerable margin, on average 4\% over the second-best, mT5. In all monolingual tasks except MTD, BanglaT5 achieves a big performance gain over others (up to 9.54\% in QA), which can be attributed to the quality of the pretraining data. In MD, BanglaT5 lags marginally behind XLM-ProphetNet. We hypothesize this is due to the lack of colloquial data in Bangla2B+ since \citet{bhattacharjee-etal-2022-banglabert} left out such sources to avoid toxic and biased conversations.

We find the MT results particularly interesting, where BanglaT5 outperforms larger multilingual models in both directions. This suggests that despite having very little English data in the pretraining corpus, BanglaT5 can generalize well to a new translation language, given high-quality fine-tuning data. We explore this more in the Appendix.
Conspicuously, all the models achieve relatively poor scores on the XLS task. This can be attributed to the smaller amount of training data.

BanglaT5 proves its superiority in compute and memory efficiency along with its performance due to its smaller size (less than half the parameters of all multilingual models except IndicBART). In practice, we observe 2-2.5x faster training and inference times with BanglaT5 than these larger multilingual models.

\section{Related Works}\label{sec:relatedworks}

\paragraph{Pretrained models} NLP has witnessed a sea of change with the advent of pretrained language models like ULMfit \citep{howard2018universal}, ELMo \citep{peters2018deep}, and most notably BERT \citep{devlin-etal-2019-bert}, achieving state-of-the-art results in many NLU benchmarks. Besides these NLU models, more and more pretrained models designed for NLG tasks have been proposed. \citet{rothe-etal-2020-leveraging} adopted pretrained NLU model checkpoints for generative tasks. GPT-2 \cite{radford2019language}, and later GPT-3 \cite{NEURIPS2020_1457c0d6} showed that pretrained generative language models can perform remarkably well in zero-shot transfer tasks. More recently, \citet{qi-etal-2020-prophetnet} proposed ProphetNet, which introduces the future n-gram prediction mechanism for language generation. \citet{dabre-etal-2022-indicbart} introduced IndicBART, which is pretrained on 11 Indic languages, including Bangla. 

\paragraph{NLG Benchmarks} Recently, many multi-task benchmarks have been proposed to drive the progress of NLG models. \citet{moussallem-etal-2020-general} proposed the BENG benchmark for NLG and knowledge extraction. GLGE \cite{liu-etal-2021-glge} is a similar benchmark with a different set of tasks and difficulty levels. However, these benchmarks are limited to English only. \citet{gehrmann-etal-2021-gem} introduced the GEM benchmark for various tasks such as summarization \cite{narayan-etal-2018-dont}, data-to-text generation \cite{nan-etal-2021-dart} across different languages. \citet{cahyawijaya-etal-2021-indonlg} introduced different tasks and baselines for 3 Indonesian languages. More recently, \citet{kumar2022indicnlg} introduced IndicNLG, a benchmark with five tasks in 11 Indic languages, including Bangla.
\section{Conclusion \& Future Works}

NLP research in low-resource languages is lagging behind due to the lack of reliable benchmarks and datasets. To facilitate the development, evaluation, and comparison of new NLG models, we introduced a multi-task evaluation benchmark for Bangla NLG, a widely spoken yet low-resource language. We presented BanglaT5, a pretrained NLG model in Bangla, setting new state-of-the-art results with BanglaT5. We strongly believe that our contributions in this work will help the Bangla NLP community benchmark NLG tasks more easily under a unified setup.

In future work, we plan to introduce new tasks to BanglaNLG, such as personalized dialogue generation \cite{zhang-etal-2018-personalizing}, conversational question-answering \cite{reddy-etal-2019-coqa}. We will also add more recent multilingual models to our comparison to BanglaT5, e.g., DeltaLM \cite{DBLP:journals/corr/abs-2106-13736}.

\section*{Limitations}

Although \citet{bhattacharjee-etal-2022-banglabert} claimed that Bangla2B+, the pretraining corpus for BanglaT5, had been carefully filtered for offensive or unwanted texts, they alerted that there might be small amounts of these contents may be present, which can result in bias or toxicity in the pretrained model. We, therefore, recommend using BanglaT5 with caution, especially for real-world deployment. 

\section*{Ethics Statement}

\paragraph{License} The TyDiQA dataset \cite{clark-etal-2020-tydi} is released under the Apache License 2.0, allowing modifications and distribution. All other pretraining and fine-tuning datasets are released under the  Creative Commons Attribution-NonCommercial-ShareAlike 4.0 International License (CC BY-NC-SA 4.0), which allows modifications and distributions for non-commercial research purposes. We strictly adhere to these licenses and will release BanglaT5 and BanglaNLG benchmark resources under CC BY-NC-SA 4.0.

\paragraph{Annotation} Expert translators who provide translation services for renowned Bangla newspapers were hired to translate the evaluation sets of the dialogue dataset. Each translated sentence was further assessed for quality by another expert. It was again translated by the original translator if found to be of low quality. If the re-translation was found to be of low quality, it was then translated by the other expert. The experts were paid hourly as per standard rates in local currency. 

\paragraph{Hallucinated Text} It is well-known that text generation models can hallucinate outputs that may not necessarily be faithful to the original input \citep{maynez-etal-2020-faithfulness}. Though the texts may be fluent and human-like, the hallucinations may be factually inconsistent and impact the outputs negatively. BanglaT5 may be susceptible to the same kinds of hallucinations.

\paragraph{Carbon Footprint} We avoided using large models for pretraining and fine-tuning, reducing their environmental impacts. BanglaT5 was trained for about 30 days on Google v3 TPUs. Google’s TPUs are specifically designed for machine learning, which makes them up to five times more efficient than GPUs. Assuming 0.080kg carbon emission per kWh,\footnote{\url{https://blog.google/technology/ai/minimizing-carbon-footprint/}} the pretraining would emit fewer than 100kg carbon into the environment, far below most computationally demanding models. All fine-tuning experiments were done on a desktop machine with an 8-core Intel Core-i7 11700k CPU and NVIDIA RTX 3090 GPU, and no single run except machine translation took more than 12 hours, which amounts to fewer than 0.5kg carbon emission. On average, machine translation runs took three days each, emitting less than 3kg of carbon.

\section*{Acknowledgements}
We would like to thank the Research and Innovation Centre for Science and Engineering (RISE), BUET, for funding the project and Google TPU Research Cloud (TRC) program for providing cloud support.

\bibliography{anthology,eacl2023}
\bibliographystyle{acl_natbib}

\clearpage
\clearpage
\appendix

\twocolumn[{%
 \centering
 \Large\bf Supplementary Material: Appendices \\ [20pt]
}]

\section{Multi-turn Dialogue Scores}

In Table \ref{tab:md}, we mention BLEU-1, BLEU-2, BLEU-3, and BLEU-4 scores for different models in the multi-turn dialogue generation task. 

\begin{table}[h]
\centering
\resizebox{\linewidth}{!}
{%
\begin{tabular}{l@{\hskip 0.05in}cccc}
\hline
Model & B-1 & B-2 & B-3 & B-4\\
\hline
mT5 (base) & 17.54 & 3.67 & 1.25 & 0.43\\
XLM-ProphetNet & \textbf{19.98} & \textbf{6.06} & \textbf{2.98} & \textbf{1.86}\\
mBART-50 & \textbf{18.54} & \textbf{5.56} & \textbf{2.97} & \textbf{2.09}\\
IndicBART (unified) & 14.75 & 3.18 & 1.06 & 0.37 \\
IndicBART (separate) & 14.05 & 3.23 & 1.18 & 0.49 \\
BanglaT5 & \textbf{19.00} & 5.02 & 2.04 & 0.92 \\
\hline
\end{tabular}
}
\caption{Performance comparison of the pretrained models on the dialogue generation task. Scores in bold texts have statistically significant ($p < 0.05$) difference from others with bootstrap sampling \citep{koehn-2004-statistical}.}
\label{tab:md}
\end{table}

\section{Cross-lingual Capabilities of BanglaT5}

Despite being a monolingual model pretrained on heavily filtered Bangla data, BanglaT5 exhibits strong cross-lingual abilities, particularly in the machine translation (MT) task. In addition to the quality and size of the fine-tuning dataset, this performance can also be attributed to the presence of a significant amount of non-Bangla tokens ($\sim$10.3\%) in the BanglaT5 vocabulary. 

Since \citet{bhattacharjee-etal-2022-banglabert} curated the Bangla2B+ corpus by document-level language filtering, these documents preserve foreign text sequences occurring in the Bangla documents. We deliberately maintain these tokens while training the vocabulary of BanglaT5, using a relatively high character coverage. Our rationale behind doing this was to capture code-switching and allow better generalization across languages co-occurring with Bangla, as well as romanized forms of Bangla texts during fine-tuning, which is reflected in the MT results. However, it should be noted that the quality and size of fine-tuning data are essential for a strong cross-lingual performance since the mere existence of foreign tokens in the vocabulary is not enough to produce meaningful generation performance, as demonstrated by the poor performance in the cross-lingual summarization (XLS) task.

This phenomenon has been studied in-depth by \citet{blevins2022language} in the context of pretrained language models in English, where they showed that these models develop strong cross-lingual transfer capabilities due to the non-negligible amount of foreign text present in the pretraining data and robustness to UNK tokens during fine-tuning.

\end{document}